\documentclass[a4paper, 10pt, conference]{ieeeconf}      % Use this line for a4
\usepackage{FG2027}
\usepackage{multirow}
\usepackage{amsmath} 
\usepackage{booktabs}
\usepackage{graphicx}
\usepackage{arydshln}
\usepackage{amssymb}
\usepackage{stfloats}
\usepackage{url}

\everypar=\expandafter{\the\everypar\loosness=-1 }
\FGfinalcopy % *** Uncomment this line for the final submission

\IEEEoverridecommandlockouts                              % This command is only
\def\FGPaperID{****} % *** Enter the FG2027 Paper ID here

\title{\LARGE \bf
Frequency-Aware Text-to-Motion Generation
}

\author{\parbox{16cm}{\centering
    {\large Dang Dinh NGUYEN$^1$, Decky ASPANDI-LATIF$^1$, Titus ZAHARIA$^1$}\\
    {\normalsize
    $^1$ SAMOVAR, Telecom SudParis, Institut Polytechnique de Paris, 91120 Palaiseau, France}}
    \thanks{This work was not supported by any organization}% <-this % stops a space
}

\begin{document}

\ifFGfinal
\thispagestyle{empty}
\pagestyle{empty}
\else
\author{Anonymous FG2027 submission\\ Paper ID \FGPaperID \\}
\pagestyle{plain}
\fi
\maketitle

%%%%%%%%%%%%%%%%%%%%%%%%%%%%%%%%%%%%%%%%%%%%%%%%%%%%%%%%%%%%%%%%%%%%%%%%%%%%%%%%
\begin{abstract}

Diffusion-based generative models have achieved significant progress in 3D human motion generation, enabling the synthesis of increasingly realistic and semantically coherent motions from natural language descriptions. Human motion exhibits rich frequency characteristics: low-frequency components capture slowly varying global motion structures, while higher-frequency components encode rapid and fine-grained motion variations. However, existing approaches primarily operate on spatio-temporal representations, leaving these characteristics largely implicit, which may limit their ability to capture the dynamics and expressive variations of human motions.  In this work, we introduce frequency information into a skeleton-structured latent diffusion framework at two stages. First, a DCT-based frequency enhancement module refines the frequency components of the skeletal-representation through learnable, body-part specific gains, shaping the latent space in which motion generation takes place. Second, a frequency attention mechanism enables the denoiser to relate temporal frequency components directly while generating in this latent space. Experiments on HumanML3D and KIT-ML demonstrate that our approach consistently improves generated motion quality over the baseline and achieves the best text-motion alignment results on HumanML3D among the other compared methods. Qualitative results further show that the proposed method better captures specific movement characteristics described by text, including rapid, pronounced motions and periodic transitions between poses. These findings highlight the potential of refining temporal frequency components to shape the latent space and exploiting them during diffusion denoising process for text-driven human motion generation.

\end{abstract}

%%%%%%%%%%%%%%%%%%%%%%%%%%%%%%%%%%%%%%%%%%%%%%%%%%%%%%%%%%%%%%%%%%%%%%%%%%%%%%%%

\section{INTRODUCTION}

% As an essential goal of artificial intelligence, generating human-like motion patterns has gained increasing interest from various communities spanning computer vision, computer graphics, robotics, and multimedia.

Modeling and generating human motion is central to how machines understand and interact with people, and has attracted increasing interest across computer vision, computer graphics, robotics, and multimedia. With the ability to synthesize natural, realistic and diverse human motions under various conditions, human motion generation can be used for a wide range of applications, including film production, video games, AR/VR, etc.~\cite{zhu2024motion,khani2025motion}. In recent years, generative paradigms - particularly diffusion models~\cite{tevet2022human, zhang2023remodiffuse, zhang2024motiondiffuse} and latent diffusion models~\cite{tevet2022motionclip, hong2025saladskeletonawarelatentdiffusion} - have made significant progress, notably in text-to-motion generation, by yielding increasingly plausible outputs.

Despite such advancements, existing approaches still present some limitations. Most frameworks model motion directly in its raw spatio-temporal form, as 3D joint positions evolving over time. In this form, properties that are essential for distinguishing fine-grained actions, such as "writing" versus "drawing", are only implicitly represented~\cite{Chang2024WaveletDecouplingCE, wu2025frequency}. Overlooking them can lead to the loss of subtle action semantics, stylistic variations, as well as physical implausibility over extended temporal horizons~\cite{wan2023diffusionphasemotiondiffusionfrequency, ZHANG20251}.

% Human motion exhibits rich frequency characteristics, with different frequency components capturing complementary aspects of movement. 

A natural way to expose these properties is to analyze motion in the frequency domain, where different components capture complementary aspects of movement: low-frequency components describe slowly varying global structures and overall locomotion, while higher-frequency components capture rapid motion variations and fine-grained dynamics. Frequency-domain analysis has shown promising results in motion recognition, retrieval, estimation, and prediction~\cite{Chang2024WaveletDecouplingCE, wu2024frequency, HE2024127495, wu2025frequency, ren2025wamo, ZHANG20251, ZHANG2025111562, wu2026khmpfrequencydomainkalmanrefinement}, but its integration into generative models remains an open research direction. Recent works have explored complementary strategies, either incorporating frequency information into the generation process or structuring the generative representation with frequency characteristics. 

% However, these approaches typically introduce frequency information at a single stage, such as the training objective, the tokenizer, or the motion representation, and treat the body as a whole rather than modeling the distinct dynamics of individual body parts.

% Inspired by the latter paradigm, we investigate how to integrate frequency information into the latent generative space while preserving the spatial organization of human motion.

% In this work, we explore how the frequency characteristics of human motion can structure the latent generative space of a diffusion model and guide generation within it. Our central idea is to integrate and refine frequency components without losing the spatial grounding essential to human motion, thereby bringing frequency and spatial representation into a unified generative framework. This perspective provides the foundation for incorporating frequency information into both latent representation learning and diffusion-based motion synthesis.

% In this work, we explore how the frequency characteristics of human motion can structure the latent space of a diffusion model and guide motion generation. Our goal is to integrate and refine frequency components while preserving the spatial structure essential to human motion. To this end, we introduce frequency modeling at two stages: first, by enhancing the motion features using frequency information to shape the latent space, and second, by incorporating frequency-aware attention into the denoising process.

In this work, we investigate how temporal frequency modeling can complement skeleton-structured latent diffusion for text-to-motion generation. We develop a framework that incorporates frequency information into both latent representation learning and the denoising process during generation, while retaining the spatial organization of human motion.

The contributions of our work are summarized as follows:

\begin{itemize}
\item We introduce a DCT-based frequency-enhancement module into the VAE encoder of a skeleton-structured latent diffusion framework, which refines temporal frequency components through learnable, body-part-specific gains, shaping the latent space while preserving its body-part organization.

\item We explore a frequency-aware attention mechanism in the diffusion denoiser, in which temporal frequency temporal bands are used as information to relate low and high frequency motion dynamics during the generation process.

\item We conduct extensive experiments on the KIT-ML~\cite{Plappert2016} and HumanML3D~\cite{Guo_2022_CVPR} dataset, where our approach achieves competitive results against state-of-the-art text-to-motion methods on both reconstruction and generation metrics.

\item We analyze the effect of frequency-enhanced latent representations on text-to-motion generation through ablation studies and joint-wise spectral analysis, showing how frequency refinement affects the dynamics of individual body parts.
\end{itemize}

% The remainder of this paper is organized as follows. Section~\ref{sec:related} reviews related work on text-to-motion generation and frequency-aware mechanisms. Section~\ref{sec:method} introduces the preliminaries, including the baseline SALAD framework, and presents the proposed methodology. Section~\ref{sec:experiments} describes the experimental setup and analyzes the results. Finally, section~\ref{sec:conclusion} concludes the paper and outlines some directions of future research.

% Section~\ref{sec:discussion} discusses the limitations of our work, and 
%%%%%%%%%%%%%%%%%%%%%%%%%%%%%%%%%%%%%%%%%%%%%%%%%%%%%%%%%%%%%%%%%%%%%%%%%%%%%%%%

\section{RELATED WORK}
\label{sec:related}

\subsection{Text-to-Motion Generation}
\label{sec:sub:related_t2m}

Driven by advances in deep generative architectures, text-driven human motion synthesis has seen rapid evolution. Early works primarily constrained generation to categorical action labels ~\cite{Guo_2020, petrovich2021actionconditioned3dhumanmotion}. Moving beyond restricted action categories, subsequent works introduced unconstrained textual conditioning by aligning cross-modal text and motion features in a shared latent space~\cite{language2pose}. Frameworks such as T2M~\cite{Guo_2022_CVPR} introduced temporal Variational Autoencoders (VAEs) to enhance generation diversity from textual prompts, while subsequent architectures integrated large-scale pretrained text encoders (e.g., CLIP~\cite{pmlr-v139-radford21a}) to strengthen semantic cross-modal alignment~\cite{TEACH:3DV:2022, temos, tevet2022motionclip}. Concurrently, discrete generative paradigms framed motion synthesis as a sequence-to-sequence problem, using Vector-Quantized VAEs (VQ-VAEs)~\cite{oord2018neuraldiscreterepresentationlearning} alongside auto-regressive Transformer backbones to generate motion tokens sequentially~\cite{TM2T, guo2024momask, jiang2024motiongpt, zhang2023generating, MotionGPT, zhong2023attt2m, zou2024parco}. Most recently, denoising diffusion models~\cite{3495724.3496298, pmlr-v37-sohl-dickstein15} have emerged as the dominant generative framework, leading a wave of diffusion-backed architectures for high-fidelity text-to-motion synthesis~\cite{chen2023executing, flame, lou2023diversemotiondiversehumanmotion, tevet2022human, fg-T2M, zhang2024motiondiffuse, zhang2023remodiffuse}. Among them, SALAD~\cite{hong2025saladskeletonawarelatentdiffusion} performs diffusion in a skeleton-structured latent space that preserves the organization of body parts, achieving strong text--motion alignment and forming the basis of our work.

\subsection{Frequency-aware Motion Generation}
\label{sec:related_workd_freq_aware}

Frequency-domain information has been explored in human motion analysis,
including recognition, retrieval, estimation, and
prediction~\cite{Chang2024WaveletDecouplingCE, wu2024frequency, HE2024127495,
wu2025frequency, ren2025wamo, ZHANG20251, ZHANG2025111562, wu2026khmpfrequencydomainkalmanrefinement}. In particular, Wu et~al.~\cite{wu2025frequency} explicitly refine the frequency content of skeleton features for zero-shot action recognition, enhancing low-frequency components and attenuating high-frequency ones through learnable weights, which inspires our frequency enhancement module. 

More recently, several studies have investigated how spectral characteristics can benefit motion generation, adopting different strategies to incorporate frequency information into motion
representations and generation processes. FTMoMamba~\cite{li2024ftmomambamotiongenerationfrequency} decomposes intermediate motion features into low- and high-frequency components to adapt the state-transition matrix of a frequency state-space model in the denoiser, complemented by a text state-space model for semantic alignment. Free-T2M~\cite{chen2025freet2mrobusttexttomotiongeneration} introduces stage-specific frequency consistency losses to address semantic planning and fine-grained motion refinement during diffusion denoising, and MotionVLA~\cite{zhang2026motionvla} adopts a dual-stream frequency-domain tokenizer that separates low-frequency pose semantics from high-frequency physical dynamics for auto-regressive humanoid motion generation. Another direction incorporates frequency structure directly into the representation used for generation. DiffusionPhase~\cite{wan2023diffusionphasemotiondiffusionfrequency} learns a compact, parameterized phase representation of motion that captures local periodicity and high-frequency details, demonstrating the potential of frequency-structured representations for generating periodic motion and smooth transitions.

Our work builds on both directions but differs in how and where frequency information is incorporated. Rather than replacing the motion representation with phase parameters, we retain a skeleton-structured latent space and refine the temporal frequency components of the encoder features that shape it, through body-part-specific adjustment. Moreover, instead of introducing frequency information solely through training losses, tokenizers, or sequence-level features, we incorporate it into both the latent representation and the denoiser through frequency-aware attention.

%%%%%%%%%%%%%%%%%%%%%%%%%%%%%%%%%%%%%%%%%%%%%%%%%%%%%%%%%%%%%%%%%%%%%%%%%%%%%%%%

\section{PROPOSED METHODOLOGY}
\label{sec:method}

% In this section, we present a frequency-aware framework for text-to-motion generation built upon SALAD~\cite{hong2025saladskeletonawarelatentdiffusion}. We first review SALAD in Sec.~\ref{sec:preliminaries}, then describe how we incorporate frequency information into its latent space (Sec.~\ref{sec:freq_enhance}) and its denoiser (Sec.~\ref{sec:freq_attn}).

% We extend the SALAD pipeline at both of its stages: reconstruction and generation. % In both stages, the ST-VAE decoder is used to map the $z$ to human motion.

\subsection{Overview}
\label{sec:overview}

\begin{figure*}[!hbtp]
\includegraphics[width=\textwidth]{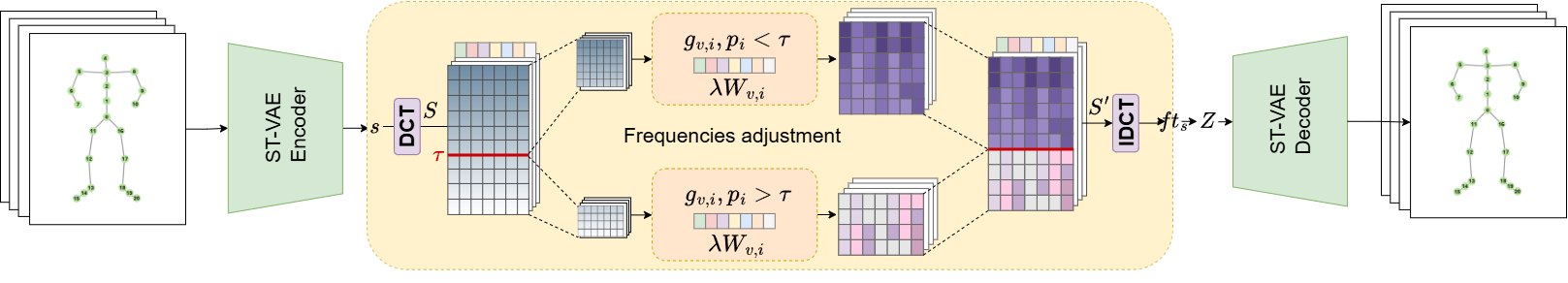}
\caption{Overview of the reconstruction pipeline that involves the DCT-based frequency enhancement module.}
\label{fig:freq_module}
%\vspace{-5mm}
\end{figure*}

\begin{figure*}[!hbt]
\includegraphics[width=\textwidth]{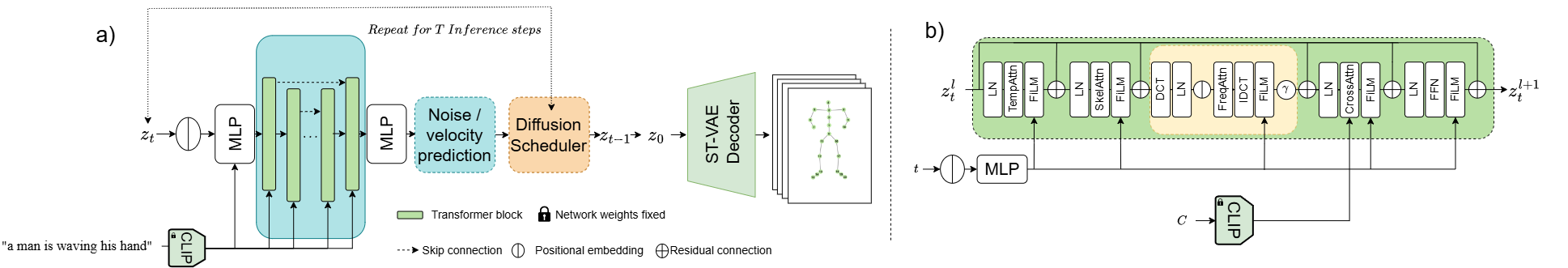}
\caption{(a)~Sampling pipeline: the denoiser iteratively refines the latent $z_t$, and the ST-VAE decoder maps the final latent $z_0$ to motion. (b)~A denoiser layer with frequency attention (highlighted), which attends across the DCT bands of each body part in the feature space.}
\label{fig:denoiser_overview}
\vspace{-5mm}
\end{figure*}

We adopt SALAD~\cite{hong2025saladskeletonawarelatentdiffusion}, a skeleton-aware latent diffusion model for text-to-motion generation, and extend it with frequency information at both its reconstruction (Fig.~\ref{fig:freq_module}) and generation (Fig.~\ref{fig:denoiser_overview}) stages.

\paragraph{Reconstruction stage - latent representation learning}
As shown in Fig.~\ref{fig:freq_module}, a spatio-temporal variational autoencoder (ST-VAE) is trained to reconstruct motion sequences. Given an input motion $\mathbf{m} \in \mathbb{R}^{F \times J \times D}$ of $F$ frames, $J$ joints and $D$-dimensional pose features, the encoder (ST-VAE encoder) maps it to features $s \in \mathbb{R}^{T \times V \times C}$ through spatio-temporal convolution and pooling, which down-sample the temporal axis to $T$ steps and group the $J$ joints into $V$ body parts, each with a $C$-dimensional representation. In our framework, these features are then refined by the frequency enhancement module (Sec.~\ref{sec:freq_enhance}), which adjusts their temporal frequency components through learnable, body-part-specific gains. The refined features $f_s$ are then mapped by fully connected heads to the $\mu$ and $\log\sigma^2$ of a Gaussian, from which the latent $z \in \mathbb{R}^{T \times V \times C}$ is sampled, and the decoder (ST-VAE decoder) reconstructs the motion $\hat{\mathbf{m}}$ from $z$. The ST-VAE, including the frequency enhancement module, is trained with
\begin{equation}
    \mathcal{L}_{\text{VAE}} = \mathcal{L}_{\text{rec}}(\hat{\mathbf{m}}, \mathbf{m})
    + w_{\text{KL}}\, \mathcal{L}_{\text{KL}},
    \label{eq:loss_vae}
\end{equation}
where $\mathcal{L}_{\text{rec}}$ is the reconstruction loss and $\mathcal{L}_{\text{KL}} = \frac{1}{2}\,\mathbb{E}\big[\mu^2 + \sigma^2 - \log\sigma^2 - 1\big]$ regularizes the latent distribution toward a standard Gaussian. Frequency information thus shapes the latent space while its body-part organization is preserved.

This structure enables the latent space as a natural basis for frequency refinement, given that different body parts often exhibit distinct frequency characteristics: the torso mainly conveys broad postural changes, whereas the limbs express more rapid, localized motion variations.

% to features $s \in \mathbb{R}^{T \times V \times C}$
% In the first stage (reconstruction) the ST-VAE encoder is used to extract features from the input skeeton sequences, then a DCT-based frequency enhancement module is applied to the encoder features before they are mapped to the latent distribution (Fig.~\ref{fig:freq_module}). It refines their temporal frequency components through learnable, body-part-specific gains, thus frequency information shapes the latent space $z$ while its body-part organization is preserved.

\paragraph{Generation stage - latent diffusion}
With the ST-VAE frozen, a transformer-based denoiser is trained to generate latents conditioned on a text description $c$, encoded by a frozen CLIP text encoder (Fig.~\ref{fig:denoiser_overview}). During training, the frequency-enhanced ST-VAE encoder maps each reference motion to its latent $z_0$, which is corrupted with noise and denoised by the network. Each denoiser layer combines temporal, skeletal, and text-motion cross-attention, to which we add a frequency attention sub-block that attends across the temporal frequency bands of each body part (Sec.~\ref{sec:freq_attn}). The denoiser $v_\theta$, including the frequency attention, is trained to predict the velocity target $v_t$ (which will be described in the next section): %(cf. Sec.~\ref{sec:overview}):
\begin{equation}
    \mathcal{L}_{\text{diff}} = \mathbb{E}_{z_0, \epsilon, t, c}
    \big[\, \| v_\theta(z_t, t, c) - v_t \|_2^2 \,\big].
    \label{eq:loss_diff}
\end{equation}

At inference, only the text is given: the denoiser iteratively refines a latent sampled from Gaussian noise, and the ST-VAE decoder maps the final latent to the output motion.

% In the second stage (Generation) the input depend on the step. During training both of the text description $c$ and the reference motions $w$ is used as the input to train the denoiser. During inference, only the text is given. The ST -VAE encoder is used to mapped the input motion to $z$,  In the denoiser, a frequency attention sub-block is added to each transformer layer, allowing the model to attend across temporal frequency bands of each body part during generation (Fig.~\ref{fig:denoiser_overview}).  with the use of DCT based frequency enhancement module  In both stages, the ST-VAE decoder is used to map the $z$ to human motion.
 
% The rest of the pipeline, including the training objective and sampling procedure, follows SALAD.

\subsection{Diffusion Formulation} 
\label{sec:preliminaries}
% SALAD is a skeleton-aware latent diffusion framework that combines a spatio-temporal variational autoencoder (ST-VAE) with a conditional diffusion denoiser. The ST-VAE encoder encodes motion sequences into a structured latent $z \in \mathbb{R}^{T \times V \times C}$, which is sampled from the approximated mean and std by the projected heads (fully conected) applied on the extracted feature of the input motion.

% in which the temporal axis is down-sampled to $T$ steps and the skeletal joints are grouped into $V$ body parts, each with its own $C$ - dimensional representation. A transformer-based denoiser then generates such latents conditioned on text $c$, the text is process through a frozen CLIP to obtain the text embeddings. then leveraging skeletal, temporal, and text--motion cross-attention we capture structural and semantic relationships. The generated latent sequences are finally decoded into 3D human motions with the ST-VAE Decoder. 

Following SALAD, we adopt a velocity-prediction parameterization~\cite{salimans2022progressive}. Given a clean motion latent $z_0$ and Gaussian noise $\epsilon \sim \mathcal{N}(0,I)$, the forward diffusion process is defined as 
\begin{equation} 
z_t = \alpha_t z_0 + \sigma_t \epsilon \end{equation}
where $\alpha_t$ and $\sigma_t$ are time-dependent coefficients controlling the contributions of the clean latent and noise, respectively. The corresponding target velocity is 
\begin{equation} 
v_t = \alpha_t \epsilon - \sigma_t z_0 \end{equation} 
Given a noisy latent $z_t$, a text condition $c$, and a timestep $t$, the denoiser predicts the velocity as
\begin{equation} 
\hat{v}_t = v_\theta(z_t,t,c) \end{equation} 
where $v_\theta$ denotes the conditional denoising network. The predicted clean latent can then be recovered as 
\begin{equation} 
\hat{z}_0 = \alpha_t z_t - \sigma_t \hat{v}_t 
\end{equation} 
% This formulation provides the basis for iterative denoising and reconstruction of motion latents. 

% Beyond providing an effective latent diffusion framework, SALAD's structured skeletal representation offers a useful foundation for body-part-specific frequency modeling. Different body parts may exhibit distinct frequency characteristics, with the torso capturing broad postural changes and the hands and feet expressing more rapid, localized variations. Exploiting this structure allows our framework to adapt frequency-aware operations to different body parts while preserving their relationships within the overall motion.

% In this section, we introduce a frequency-aware framework for text-to-motion generation built upon SALAD~\cite{hong2025saladskeletonawarelatentdiffusion}. Our approach first employs a DCT-based frequency enhancement module to adaptively modulate the frequency components of motion latents, producing a frequency-structured latent representation that captures both global motion patterns and fine-grained temporal variations while retaining the spatial organization of body parts. This enhanced representation is then used within the diffusion process, where a frequency-aware attention mechanism enables the denoiser to exploit frequency information during motion synthesis. 

% By integrating frequency adjustment into both latent representation learning and denoising, our framework combines structured motion representations with frequency-aware generation. 

\subsection{DCT-Based Frequency Enhancement}
\label{sec:freq_enhance}

Inspired by the frequency-adjustment strategy of~\cite{wu2025frequency}, we introduce a Discrete Cosine Transform (DCT)-based frequency-enhancement module into SALAD's structured latent representation. The module modulates temporal frequency components through learnable, body-part-specific gains, allowing each body-part group to adapt its latent representation to distinct motion dynamics. To accommodate variations in sequence length, the modulation is defined over normalized frequency positions rather than absolute coefficient indices. 

% The overall architecture is illustrated in Fig.~\ref{fig:freq_module}.

% The encoder features $s \in \mathbb{R}^{T \times V \times C}$ are transformed by the DCT along the temporal axis, and the resulting coefficients $\mathcal{S}$ are grouped into $B$ frequency bands for each of the $V$ body parts. Bands below the threshold $\tau$ are enhanced and bands above it are attenuated by body-part-specific gains $g_{v,i}$, parameterized by $\lambda\mathbf{W}_{v,i}$. The modulated coefficients $\mathcal{S}'$ are mapped back by the IDCT to obtain the enhanced features $f_s$.

Given the ST-VAE encoder features $s \in \mathbb{R}^{T \times V \times C}$, we apply the DCT method along the temporal dimension to extract the frequency coefficients $\mathcal{S}$:
\begin{equation}
    \mathcal{S}_{k,v,c}
    = \alpha_k \sum_{t=0}^{T-1} s_{t,v,c}
    \cos\left(\frac{\pi(2t+1)k}{2T}\right),
    \label{eq:dct}
\end{equation}
where $k \in \{0,\ldots,T-1\}$ is the frequency index, $\alpha_0 = 1/\sqrt{T}$, and $\alpha_k = \sqrt{2/T}$ for $k > 0$, $v \in V$, $c\in C$. Lower-frequency coefficients describe slowly varying temporal patterns, whereas higher-frequency coefficients capture rapid variations and fine-grained dynamics.

The coefficients are grouped into a discrete set of $B$ frequency bands, each band $i$ being represented by its normalized position $p_i = i/B$. We learn body-part-specific weights $\mathbf{W} \in \mathbb{R}^{V \times B}$ that adjust the contribution of each band through a gain function \textit{g} that enhances the lower-frequency region and attenuates the higher-frequency region, defined as:
\begin{equation}
    g_{v,i} =
    \begin{cases}
        1 + \lambda W_{v,i}\left(1 - \dfrac{p_i}{\zeta}\right), & p_i < \tau, \\[1.5ex]
        1 - \lambda W_{v,i}\left(1 - \dfrac{p_i-\zeta}{\zeta}\right), & p_i \ge \tau,
    \end{cases}
    \label{eq:band_gain}
\end{equation}
where $\tau$ is the frequency threshold, $\zeta$ controls the frequency-dependent decay, and $\lambda$ controls the modulation magnitude. In our implementation, the threshold can be learned, and a differentiable approximation around the threshold facilitates gradient-based optimization.

% We also clamp the gains from below at $g_{\min}$ to prevent excessively small scaling factors.

Each coefficient is then scaled by the gain of the band $b(k)$ containing it. Lastly, the enhanced feature is recovered with the inverse DCT:
\begin{equation}
    ft_s = \operatorname{IDCT}\left(\mathcal{S}'\right), \qquad
    \mathcal{S}'_{k,v,c} = g_{v,b(k)} .\, \mathcal{S}_{k,v,c}.
    \label{eq:idct}
\end{equation}
The output $ft_s$ has the same shape as $s$, then pass through the projection head to produce $z$, preserving SALAD's structured latent layout. This module provides a lightweight mechanism for adjusting temporal frequency contributions according to body-part-specific motion characteristics, without assuming that all high-frequency components are irrelevant.

%%%%%%%%%%%%%%%%%%%%%%%%%%%%%%%%%%%%%%%%%%%%%%%%%%%%%%%%%%%%%%%%%%%%%%%%%%%%%%%%

\subsection{Frequency Attention}
\label{sec:freq_attn}

The frequency enhancement module incorporates frequency information into the latent representation, but the denoiser accesses this information only implicitly through the latent. Within SALAD's denoiser, temporal and skeletal attention both operate in the temporal domain, where the global structure of a motion, such as its overall rhythm and slowly varying trajectory, is spread across all frames. To make frequency characteristics directly available during denoising, we add a frequency attention sub-block in which the tokens are temporal frequency bands rather than time steps, allowing the denoiser to relate coarse and fine motion components explicitly.

Let $\mathbf{h}_v \in \mathbb{R}^{L \times D_h}$ denote the hidden features of body part $v$ in a sequence with $L$ valid frames. We transform them with the orthonormal DCT matrix $\mathbf{O}_L \in \mathbb{R}^{L \times L}$, constructed for the valid length of each sample, therefore padded frames do not affect the spectrum:
\begin{equation} 
    \mathbf{H}_v = \mathbf{O}_L \, \mathbf{h}_v . 
    \label{eq:fa_dct} 
\end{equation}
Each row of $\mathbf{H}_v$ is a $D_h$-dimensional token describing one frequency band. After layer normalization, we add a sinusoidal embedding $\mathbf{P}$ of the band index and apply multi-head self-attention (MHSA) across the $L$ bands, then map the result back to the time domain:
\begin{equation}
    \tilde{\mathbf{h}}_v = \mathbf{O}_L^{\top}\,
    \operatorname{MHSA}\!\big(\operatorname{LN}(\mathbf{H}_v) + \mathbf{P}\big).
    \label{eq:fa_attn}
\end{equation}
Each band can thus attend to all other bands of the same body part, with the attention weights shared across body parts. Since the number of bands equals the number of frames, the cost matches that of the temporal attention.

As for the temporal and skeletal attention, the output is modulated by the diffusion timestep $t$ through feature-wise linear modulation (FiLM)~\cite{perez2018film}, and it is added to the residual stream with a learnable scale $\gamma$:
\begin{equation}
    \mathbf{h}_v \leftarrow \mathbf{h}_v
    + \gamma \cdot \operatorname{FiLM}\big(\tilde{\mathbf{h}}_v;\, t\big).
    \label{eq:fa_res}
\end{equation}
We initialize $\gamma$ to $10^{-3}$ and the FiLM parameters to the identity, so the denoiser initially behaves as the original SALAD and gradually learns how much frequency-domain context to incorporate. In every transformer layer, the frequency attention follows the temporal and skeletal attention and precedes the cross-attention to the text.

%%%%%%%%%%%%%%%%%%%%%%%%%%%%%%%%%%%%%%%%%%%%%%%%%%%%%%%%%%%%%%%%%%%%%%%%%%%%%%%%

\section{EXPERIMENTS}
\label{sec:experiments}

\subsection{Experimental Settings} 

\textbf{Dataset} We evaluate our approach on two widely used text-to-motion benchmarks, HumanML3D \cite{Guo_2022_CVPR} and KIT-ML \cite{Plappert2016}. HumanML3D contains 14,616 motion sequences covering a wide range of human activities, paired with 44,970 textual descriptions. KIT-ML contains 3,911 motion sequences and 6,278 textual descriptions. Both datasets are augmented by motion mirroring and divided into training, validation, and test sets using a ratio of 0.80, 0.05, and 0.15, respectively.

\textbf{Metrics} We evaluate the model at two levels: \textbf{motion reconstruction} and \textbf{motion generation}. For the reconstruction capability, we use Mean Per Joint Position Error (MPJPE) \cite{ahuja2019language2pose} and Fréchet Inception Distance (FID) \cite{fid}. MPJPE measures the positional discrepancy between reconstructed and ground-truth motions, while FID evaluates the similarity between their feature distributions.

For the text to motion generation evaluation, we use FID metric to assess the quality of generated motions and R-precision and MM-Dist \cite{Guo_2022_CVPR} metric to measure the semantic alignment between generated motions and their corresponding text descriptions. We also report Diversity and MultiModality \cite{lee2019dancing} score to evaluate the variability of generated motions across different text conditions and the diversity of motions generated from the same text condition, respectively. Each experiment is repeated 20 times on the test sets, and we report the mean performance with a 95\% confidence interval.

All metrics are computed with the pretrained motion and text feature extractors of~\cite{Guo_2022_CVPR}. As a reference, we also report the metrics obtained on the ground-truth test motions, denoted as Real Motion, which indicate the values attainable by real data under the same feature extractors. For the compared methods, we report the results published in their original papers. Entries marked with ``-'' are not reported by the corresponding work.

% to ensure a fair comparison with the baseline

\textbf{Implementation} Our method baseline is built upon SALAD base code ~\cite{hong2025saladskeletonawarelatentdiffusion}. All experiments have been conducted using two NVIDIA RTX 4090 GPUs. We used the AdamW optimizer~\cite{loshchilov2019decoupledweightdecayregularization} to train the VAE and denoiser for 50 and 500 epochs, respectively. The batch size was set to 64 for HumanML3D and 16 for KIT-ML. The diffusion model was trained with 1,000 timesteps, while inference employed 50-step DDIM sampling. The classifier-free guidance (CFG) scale was set to (w=7.5). \footnote{The implementation of our method is publicly available in our repository \url{https://anonymous.4open.science/r/Frequency-Aware_T2M-0664/}}

%%%%%%%%%%%%%%%%%%%%%%%%%%%%%%%%%%%%%%%%%%%%%%%%%%%%%%%%%%%%%%%%%%%%%%%%%%%%%%%%

\subsection{Comparison with the SALAD Baseline}
\label{sec:baseline}
\subsubsection{Effectiveness of frequency-aware Latent Space}

\begin{table}[!htbp] 
\centering
\caption{Quantitative results of reconstructed motion quality on HumanML3D and KIT-ML dataset. $\downarrow$ denote lower values are preferable. \textbf{Bold} text indicate best results.}
\label{tab:reconstruction_results}
% \vspace{-1mm}
\setlength{\tabcolsep}{2pt} % Reduce inter-column spacing naturally
\fontsize{7pt}{9pt}\selectfont
\begin{tabular}{l c c c c}
\toprule
 & \multicolumn{2}{c}{\textbf{HumanML3D}} & \multicolumn{2}{c}{\textbf{KIT-ML}} \\
\cmidrule(lr){2-3} \cmidrule(lr){4-5}
\textbf{Method} & FID $\downarrow$ & MPJPE $\downarrow$ & FID $\downarrow$ & MPJPE $\downarrow$ \\ 
\midrule
SALAD & \textbf{0.003 $\mathbf{\pm}$ 0.000} & \textbf{0.016 $\mathbf{\pm}$ 0.000} & $0.074 \pm 0.002$ & $15.167 \pm 0.018$ \\
\midrule
\textbf{Ours} & 0.009 $\pm$ 0.000 & \textbf{0.016 $\pm$ 0.000} & \textbf{0.055 $\mathbf{\pm}$ 0.001} & \textbf{14.957 $\mathbf{\pm}$ 0.087} \\
\bottomrule
\end{tabular}
%\vspace{-3mm}
\end{table}

To assess how frequency affects representation capacity, we evaluate the reconstruction quality of the underlying VAE models in both SALAD baseline and ours on HumanML3D and KIT-ML (Table~\ref{tab:reconstruction_results}). On HumanML3D, our method preserves near-lossless reconstruction fidelity, maintaining an MPJPE of 0.016 and an FID of $0.006$, which is very close to real motion ($0.002$). On KIT-ML, introducing temporal frequency modulation yields substantial reconstruction gains over the baseline SALAD model as it lowers the reconstruction FID from $0.074$ to $0.055$ (approaching the ground-truth benchmark of $0.031$) while reducing MPJPE from $15.167$ to $14.957$. These results confirm that explicit frequency adjustment acts as an effective regularizer, filtering non-essential variance while maintaining and enhancing the spatial-temporal reconstruction capabilities of the autoencoder on challenging datasets.

\begin{table*}[!htbp]
\caption{Quantitative comparison of SALAD and our method on the HumanML3D and KIT-ML test sets. $\uparrow$ and $\downarrow$ indicates that higher and lower values are better, respectively. $\rightarrow$ indicates that values closer to those of real motions are preferable.}
\label{tab:baseline_comparison}
\centering
\resizebox{\textwidth}{!}{%
\begin{tabular}{l l ccc c c c c}
\toprule
\multirow{2}{*}{Dataset} & \multirow{2}{*}{Method} & \multicolumn{3}{c}{R-Precision $\uparrow$} & \multirow{2}{*}{FID $\downarrow$} & \multirow{2}{*}{MM-Dist $\downarrow$} & \multirow{2}{*}{Diversity $\rightarrow$} & \multirow{2}{*}{MultiModality $\uparrow$} \\
\cmidrule(lr){3-5}
& & Top-1 & Top-2 & Top-3 & & & & \\
\midrule
\multirow{3}{*}{\textbf{HumanML3D}}
& Real Motion & $0.511 \pm 0.003$ & $0.703 \pm 0.003$ & $0.797 \pm 0.002$ & $0.002 \pm 0.000$ & $2.974 \pm 0.008$ & $9.503 \pm 0.065$ & -- \\
\cline{2-9}
& SALAD & $0.581 \pm 0.003$ & $0.769 \pm 0.003$ & $0.857 \pm 0.002$ & $0.076 \pm 0.002$ & $2.649 \pm 0.009$ & \textbf{9.696 $\pm$ 0.096} & \textbf{1.751 $\pm$ 0.062} \\
& Ours & \textbf{0.585 $\pm$ 0.005} & \textbf{0.781 $\pm$ 0.004} & \textbf{0.868 $\pm$ 0.003} & \textbf{0.052 $\pm$ 0.003} & \textbf{2.473 $\pm$ 0.08} & $9.926 \pm 0.066$ & $1.463 \pm 0.065$ \\
\midrule
\multirow{3}{*}{\textbf{KIT-ML}}
& Real Motion & $0.424 \pm 0.005$ & $0.649 \pm 0.006$ & $0.779 \pm 0.006$ & $0.031 \pm 0.004$ & $2.788 \pm 0.012$ & $11.080 \pm 0.097$ & -- \\
\cline{2-9}
& SALAD & \textbf{0.477 $\pm$ 0.006} & \textbf{0.711 $\pm$ 0.005} & \textbf{0.828 $\pm$ 0.005} & $0.296 \pm 0.012$ & $2.585 \pm 0.016$ & \textbf{11.097 $\pm$ 0.095} & $1.004 \pm 0.040$ \\
& Ours & $0.465 \pm 0.005$ & $0.690 \pm 0.007$ & $0.809 \pm 0.006$ & \textbf{0.182 $\pm$ 0.010} & \textbf{2.575 $\pm$ 0.013} & $11.124 \pm 0.102$ & \textbf{1.013 $\pm$ 0.049} \\
\bottomrule
\end{tabular}%
}
\vspace{-5mm}
\end{table*}
Table~\ref{tab:baseline_comparison} compares our method with the SALAD baseline the on generation task. The most consistent improvement is in motion quality: FID decreases from $0.076$ to $0.052$ on HumanML3D and from $0.296$ to $0.182$ on KIT-ML, which corresponds to relative reductions of $32\%$ and $38\%$, respectively. On HumanML3D, our method also improves text alignment, with higher R-precision at all three ranks and a lower MM-Dist ($2.473$ vs.\ $2.649$). On KIT-ML, MM-Dist results are on par with the results from SALAD, while R-precision decreases slightly. Diversity remains close to that of real motions on both datasets, and MultiModality is comparable on KIT-ML, whereas on HumanML3D both metrics are slightly less favorable than for SALAD. This may stem from differences in the data: KIT-ML is considerably smaller than HumanML3D and differs in coordinate scale, which may affect the statistics of the learned features, and its motions may contain more high-frequency noise, which could reduce the benefit of our frequency enhancement. Overall, these results indicate that incorporating frequency information can improve the fidelity of the generated motions, at a modest cost in generation variability.

\subsubsection{Qualitative Results}
\label{sec:sub:sub:baseline_qualitative}
\begin{figure*}[!htbp]
\centering
\includegraphics[width=\textwidth]{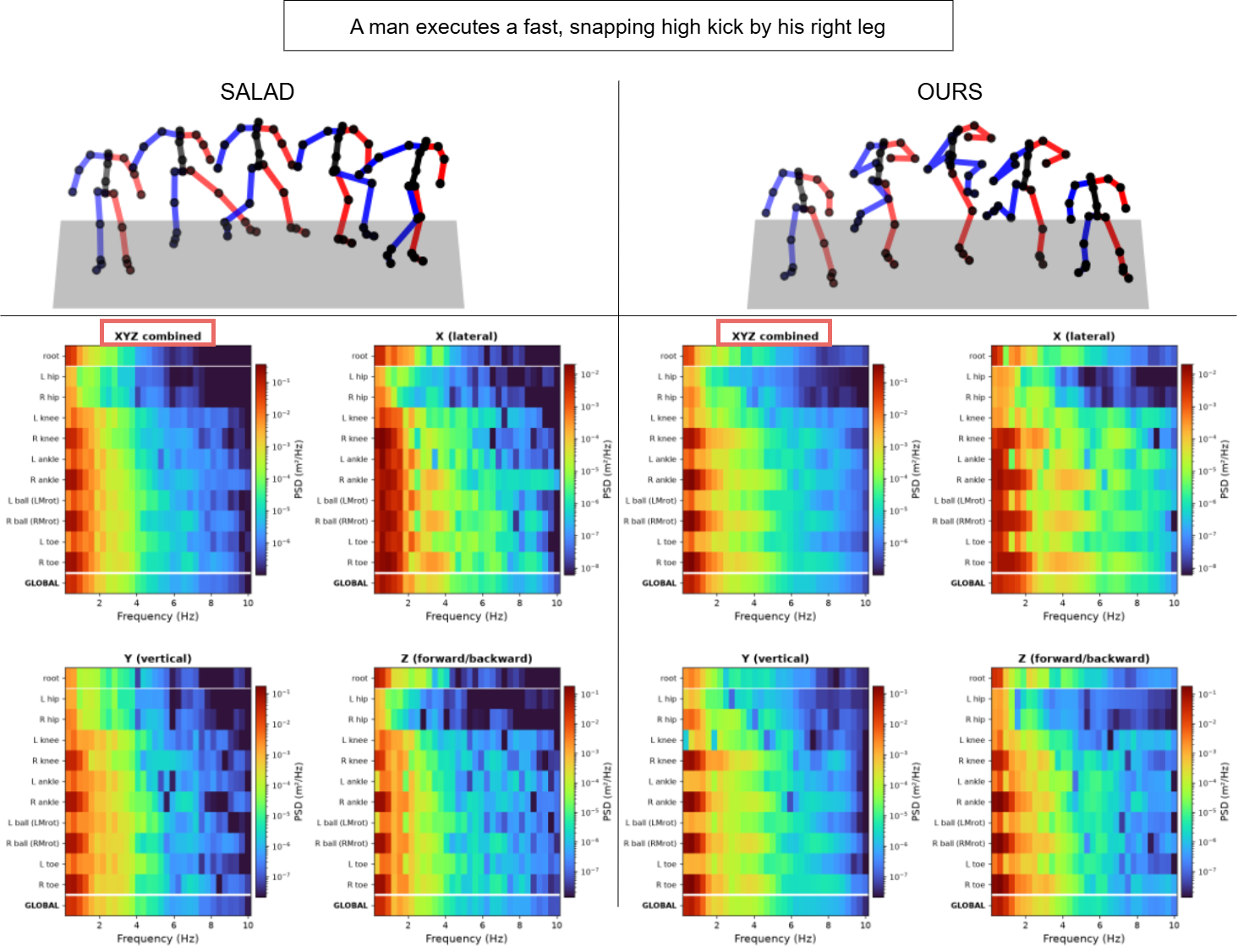}
\caption{Qualitative and frequency comparison between SALAD (left) and our method (right) for the prompt of ``A person executes a fast, snapping high kick by his right leg.'' Top: generated motions. Bottom: joint-wise power spectra of the generated motions (combined XYZ and per axis)}
% , with the GLOBAL row summarizing all joints.
\label{fig:salad_qualitative}
\vspace{-5mm}
\end{figure*}
Figure~\ref{fig:salad_qualitative} presents a qualitative comparison between SALAD and our method for the prompt of ``A person executes a fast, snapping high kick by his right leg.'' In the example of the generated motions (top row), the motion generated by SALAD begins with an additional displacement to the left before performing the kick, introducing movement that is not described in the text. Its kicking trajectory also appears less clear. In contrast, our method produces a more focused progression toward the target action, with a clearer and more pronounced leg movement. These observations suggest that our method better captures the motion dynamics described in the prompt.

The joint-wise frequency-power distributions (bottom row of the figure) provide further insight into these result differences. Compared with SALAD, our method change the distribution of spectral power across joints and movement directions. Along the vertical axis, we can see the increase of intensity for the right-leg joints (such as the ankle, knee and foot) and extends into higher frequency bins, reflecting both the large amplitude and the rapid changes of the kick; it also shows clear distinction of the power between right-leg (high intensity, relevant) and left-leg (less intensity, thus is not relevant) joints showing the semantic alignment with the provided text prompt. This pattern is nearly absent for SALAD,  which instead concentrates lateral power (x axes) on the left (supporting) leg, extending into the higher bins, which is consistent with the sideways displacement visible in its motion (thus focus is instead in the moving of the body). These results suggest that our frequency modeling directs spectral power toward the body part and the dynamics specified by the text.

%Our method shows a modest increase in high-frequency power in the torso, which we attribute partly to the body’s response to the kick. 

%%%%%%%%%%%%%%%%%%%%%%%%%%%%%%%%%%%%%%%%%%%%%%%%%%%%%%%%%%%%%%%%%%%%%%%%%%%%%%%%

\subsection{Comparison with State-of-the-Art Methods}

We compare our method with approaches from two generative paradigms. Among auto-regressive and token-based methods, T2M~\cite{Guo_2022_CVPR} generates motion snippets sequentially with a temporal VAE; AttT2M~\cite{zhong2023attt2m} generates discrete tokens with body-part and text-motion attention; ParCo~\cite{zou2024parco} coordinates separate body-part generators; and MoMask~\cite{guo2024momask}predicts residual-quantized tokens with masked Transformers. Among diffusion-based methods, MDM~\cite{tevet2022human} denoises raw motion with a Transformer; MLD~\cite{chen2023executing} moves denoising into a VAE latent space; MotionDiffuse~\cite{zhang2024motiondiffuse} enables body-part and time-specific control through cross-modal attention; and ReMoDiffuse~\cite{zhang2023remodiffuse} guides denoising with retrieved reference motions. We further compare with frequency-aware methods, namely DiffusionPhase~\cite{wan2023diffusionphasemotiondiffusionfrequency}, FTMoMamba~\cite{li2024ftmomambamotiongenerationfrequency}, Free-T2M~\cite{chen2025freet2mrobusttexttomotiongeneration}, and MotionVLA~\cite{zhang2026motionvla}.

% , and with our baseline SALAD~\cite{hong2025saladskeletonawarelatentdiffusion}.

\subsubsection{Quantitative Evaluations}

% Methods above the first dotted line are auto-regressive models, while those below are diffusion-based generative models.

\begin{table}[!htbp] 
\centering
\caption{Quantitative results on the reconstructed motion quality on HumanML3D and KIT-ML dataset. \textbf{Bold} and \textcolor{blue}{blue} indicate best and second best results, respectively.}
\label{tab:reconstruction_sota_results}
% \vspace{-1mm}
\setlength{\tabcolsep}{2pt} % Reduce inter-column spacing naturally
\fontsize{7pt}{9pt}\selectfont
\begin{tabular}{l c c c c}
\toprule
 & \multicolumn{2}{c}{\textbf{HumanML3D}} & \multicolumn{2}{c}{\textbf{KIT-ML}} \\
\cmidrule(lr){2-3} \cmidrule(lr){4-5}
\textbf{Method} & FID $\downarrow$ & MPJPE $\downarrow$ & FID $\downarrow$ & MPJPE $\downarrow$ \\ 
\midrule
ParCo & $0.021 \pm 0.001$ & $0.108 \pm 0.000$ & $0.311 \pm 0.006$ & — \\
MoMask & $0.019 \pm 0.001$ & $0.029 \pm 0.000$ & $0.112 \pm 0.002$ & $37.500 \pm 0.000$ \\
SALAD & \textbf{0.003 $\mathbf{\pm}$ 0.000} & \textbf{0.016 $\mathbf{\pm}$ 0.000} & \textcolor{blue}{$0.074 \pm 0.002$} & \textcolor{blue}{$15.167 \pm 0.018$} \\
\midrule
\textbf{Ours} & \textcolor{blue}{$0.009 \pm 0.000$} & \textbf{0.016 $\pm$ 0.000} & \textbf{0.055 $\mathbf{\pm}$ 0.001} & \textbf{14.957 $\mathbf{\pm}$ 0.087} \\
\bottomrule
\end{tabular}
% \vspace{-3mm}
\end{table}

Seeing the results from the Table~\ref{tab:reconstruction_sota_results} for the reconstruction task, our results achieved the best results overall in comparison with other methods such as ParCo and MoMask highlighting ability of our methods to reconstruct the motions input.

\begin{table*}[!htbp] 
\caption{Quantitative evaluation results on the test sets of HumanML3D (top) and KIT-ML (bottom). $\uparrow$ and $\downarrow$ denote that higher and lower values are better, respectively, while $\rightarrow$ denotes that the values closer to the real motion are better. \textbf{Bold}, \textcolor{blue}{blue}, and \textcolor{red}{red} text indicate the best, second-best, and third-best results, respectively.}
\label{tab:generation_results}
% \vspace{-3mm}
\centering
\resizebox{\textwidth}{!}{%
\begin{tabular}{l l ccc c c c c}
\toprule
\multirow{2}{*}{Dataset} & \multirow{2}{*}{Method} & \multicolumn{3}{c}{R-Precision $\uparrow$} & \multirow{2}{*}{FID $\downarrow$} & \multirow{2}{*}{MM-Dist $\downarrow$} & \multirow{2}{*}{Diversity $\rightarrow$} & \multirow{2}{*}{MultiModality $\uparrow$} \\
\cmidrule(lr){3-5}
 & & Top-1 & Top-2 & Top-3 & & & & \\ 
\midrule
\multirow{12}{*}{\rotatebox[origin=c]{90}{\textbf{HumanML3D}}} 
 & Real Motion & $0.511 \pm 0.003$ & $0.703 \pm 0.003$ & $0.797 \pm 0.002$ & $0.002 \pm 0.000$ & $2.974 \pm 0.008$ & $9.503 \pm 0.065$ & - \\ 
\cline{2-9}
 & T2M & $0.457 \pm 0.002$ & $0.639 \pm 0.003$ & $0.740 \pm 0.003$ & $1.067 \pm 0.002$ & $3.340 \pm 0.008$ & $9.188 \pm 0.002$ & $2.090 \pm 0.083$ \\
 & AttT2M & $0.499 \pm 0.003$ & $0.690 \pm 0.002$ & $0.786 \pm 0.002$ & $0.112 \pm 0.006$ & $3.038 \pm 0.007$ & $9.700 \pm 0.090$ & $2.452 \pm 0.051$ \\
 & ParCo & $0.515 \pm 0.003$ & $0.706 \pm 0.007$ & $0.801 \pm 0.002$ & $0.109 \pm 0.005$ & $2.927 \pm 0.008$ & \textcolor{red}{$9.576 \pm 0.088$} & $1.382 \pm 0.060$ \\
 & MoMask & $0.521 \pm 0.002$ & $0.713 \pm 0.002$ & $0.807 \pm 0.002$ & \textbf{0.045 $\mathbf{\pm}$ 0.002} & $2.958 \pm 0.008$ & - & $1.241 \pm 0.040$ \\

% \cdashline{2-9}

 & MDM & $0.320 \pm 0.005$ & $0.498 \pm 0.004$ & $0.611 \pm 0.007$ & $0.544 \pm 0.044$ & $5.556 \pm 0.027$ & \textcolor{blue}{9.559 $\mathbf{\pm}$ 0.086} & \textcolor{red}{2.799 $\mathbf{\pm}$ 0.072} \\
 & MLD & $0.481 \pm 0.003$ & $0.673 \pm 0.003$ & $0.772 \pm 0.002$ & $0.473 \pm 0.013$ & $3.196 \pm 0.010$ & $9.724 \pm 0.082$ & $2.413 \pm 0.079$ \\
 & MotionDiffuse & $0.491 \pm 0.001$ & $0.681 \pm 0.001$ & $0.782 \pm 0.002$ & $0.630 \pm 0.001$ & $3.113 \pm 0.001$ & $9.410 \pm 0.049$ & $1.553 \pm 0.042$ \\
 & ReMoDiffuse & $0.510 \pm 0.005$ & $0.698 \pm 0.006$ & $0.795 \pm 0.004$ & $0.103 \pm 0.004$ & $2.974 \pm 0.016$ & $9.018 \pm 0.075$ & $1.795 \pm 0.043$ \\
 & DiffusionPhase & - & - & $0.761$ & $0.397$ & $3.194$ & $9.363$ & \textcolor{blue}{$2.805$} \\
 & Free-T2M & \textcolor{red}{$0.555 \pm 0.003$ } & \textcolor{red}{$0.752 \pm 0.003$} & \textcolor{red}{$0.841 \pm 0.002$} & \textcolor{red}{$0.060 \pm 0.004$} & \textcolor{red}{$2.749 \pm 0.011$} & $9.710 \pm 0.138$ & $1.634 \pm 0.102$ \\
& FTMoMamba & $0.489 \pm 0.003$ & $0.680 \pm 0.002$ & $0.777 \pm 0.002$ & $0.181 \pm 0.009$ & $3.151 \pm 0.009$ & $9.789 \pm 0.085$ & $2.277 \pm 0.099$ \\
 & MotionVLA & $0.507 \pm 0.002$ & $0.699 \pm 0.002$ & $0.798 \pm 0.002$ & $0.071 \pm 0.003$ & $2.906 \pm 0.009$ & \textbf{9.548 $\pm$ 0.081} & \textbf{2.821 $\pm$ 0.091} \\
 
\cline{2-9}

& SALAD & \textcolor{blue}{0.581 $\mathbf{\pm}$ 0.003} & \textcolor{blue}{$0.769 \pm 0.003$} & \textcolor{blue}{$0.857 \pm 0.002$} & $0.076 \pm 0.002$ & \textcolor{blue}{$2.649 \pm 0.009$} & $9.696 \pm 0.096$ & $1.751 \pm 0.062$ \\
& Ours & \textbf{0.585 $\pm$ 0.005} & \textbf{0.781 $\pm$ 0.004} & \textbf{0.868 $\pm$ 0.003} & \textcolor{blue}{0.052 $\pm$ 0.003} & \textbf{2.473 $\pm$ 0.008} & $9.926 \pm 0.066$ & $1.463 \pm 0.065$ \\

\midrule \midrule

\multirow{12}{*}{\rotatebox[origin=c]{90}{\textbf{KIT-ML}}} 
 & Real Motion & $0.424 \pm 0.005$ & $0.649 \pm 0.006$ & $0.779 \pm 0.006$ & $0.031 \pm 0.004$ & $2.788 \pm 0.012$ & $11.080 \pm 0.097$ & - \\ 
\cline{2-9}
 & T2M & $0.370 \pm 0.005$ & $0.569 \pm 0.007$ & $0.693 \pm 0.006$ & $2.770 \pm 0.109$ & $3.401 \pm 0.008$ & $10.910 \pm 0.119$ & $1.482 \pm 0.065$ \\
 & AttT2M & $0.413 \pm 0.006$ & $0.632 \pm 0.006$ & $0.751 \pm 0.006$ & $0.870 \pm 0.039$ & $3.039 \pm 0.021$ & $10.960 \pm 0.123$ & \textbf{2.281 $\mathbf{\pm}$ 0.047} \\
 & ParCo & $0.430 \pm 0.004$ & $0.649 \pm 0.007$ & $0.772 \pm 0.006$ & $0.453 \pm 0.027$ & $2.820 \pm 0.028$ & $10.950 \pm 0.094$ & $1.245 \pm 0.022$ \\
 & MoMask & $0.433 \pm 0.007$ & $0.656 \pm 0.005$ & $0.781 \pm 0.005$ & 0.204 $\mathbf{\pm}$ 0.011 & $2.779 \pm 0.022$ & - & $1.131 \pm 0.043$ \\
 
% \cdashline{2-9}
 
 & MDM & $0.164 \pm 0.004$ & $0.291 \pm 0.004$ & $0.396 \pm 0.004$ & $0.497 \pm 0.021$ & $9.191 \pm 0.022$ & $10.847 \pm 0.109$ & $1.907 \pm 0.214$ \\
 & MLD & $0.390 \pm 0.008$ & $0.609 \pm 0.008$ & $0.734 \pm 0.007$ & $0.404 \pm 0.027$ & $3.204 \pm 0.027$ & $10.800 \pm 0.117$ & \textcolor{blue}{$2.192 \pm 0.071$} \\
 & MotionDiffuse & $0.417 \pm 0.004$ & $0.621 \pm 0.004$ & $0.739 \pm 0.004$ & $1.954 \pm 0.062$ & $2.958 \pm 0.005$ & \textcolor{blue}{$11.100 \pm 0.143$} & $0.730 \pm 0.013$ \\
 & ReMoDiffuse & $0.427 \pm 0.014$ & $0.641 \pm 0.004$ & $0.765 \pm 0.055$ & \textcolor{blue}{$0.155 \pm 0.006$} & $2.814 \pm 0.012$ & $10.800 \pm 0.105$ & $1.239 \pm 0.028$ \\
 & DiffusionPhase & - & - & $0.736$ & $0.503$ & $3.167$ & $10.873$ & \textcolor{red}{$2.071$} \\
 & Free-T2M & \textcolor{red}{$0.454 \pm 0.007$} & \textcolor{red}{$0.675 \pm 0.005$} & \textcolor{red}{$0.798 \pm 0.007$} & \textbf{0.150 $\pm$ 0.019} & \textcolor{red}{$2.689 \pm 0.020$} & $10.977 \pm 0.102$ & $1.141 \pm 0.064$ \\
\cline{2-9}
  & SALAD & \textbf{0.477 $\mathbf{\pm}$ 0.006} & \textbf{0.711 $\mathbf{\pm}$ 0.005} & \textbf{0.828 $\mathbf{\pm}$ 0.005} & $0.296 \pm 0.012$ & \textcolor{blue}{$2.585 \pm 0.016$} & \textbf{11.097 $\mathbf{\pm}$ 0.095} & $1.004 \pm 0.040$ \\
 & Ours & \textcolor{blue}{$0.465 \pm 0.005$} & \textcolor{blue}{$0.690 \pm 0.007$} & \textcolor{blue}{$0.809 \pm 0.006$} & \textcolor{red}{0.182 $\pm$ 0.010} & \textbf{2.575 $\pm$ 0.013} & \textcolor{red}{$11.124 \pm 0.102$} & 1.013 $\pm$ 0.049 \\
\bottomrule
\end{tabular}
}
\vspace{-5mm}
\end{table*}

Table~\ref{tab:generation_results} compares our method with state-of-the-art text-to-motion approaches. Our method achieves the best overall balance between motion quality and text alignment. On HumanML3D, it is the only method that ranks in the top two on all fidelity and alignment metrics: it achieves the highest R-precision at all three ranks (Top-1 of 0.585) and the lowest MM-Dist (2.473), and its FID of 0.052 is second only to MoMask, whose R-precision is considerably lower (Top-3 of 0.807 vs.\ 0.868). 

On KIT-ML, it obtains the lowest MM-Dist (2.575), the second-best R-precision, and the third-best FID. A plausible explanation for the strong alignment is that text descriptions often specify the temporal character of an action, such as its rhythm, repetition, or speed, and these properties are compactly reflected in the frequency domain. By exposing frequency information both in the latent space and in the denoiser, our method can match them more directly, which is consistent with the improved text alignment over our baseline and with the qualitative and spectral analyses (as we seen in Sec.~\ref{sec:sub:sub:baseline_qualitative}). MultiModality remains lower than that of methods such as MDM and MotionVLA, which obtain higher variability at a considerable cost in fidelity and alignment. 

%The improvements over our baseline are analyzed in Sec.~\ref{sec:baseline}.

Our method also outperforms frequency-aware approaches. It achieves higher R-precision and lower MM-Dist than DiffusionPhase, FTMoMamba, MotionVLA, and Free-T2M on all reported datasets, and the lowest FID among them on HumanML3D. The strongest of these, Free-T2M, introduces frequency notion through training losses; by incorporating this information into both the latent space and the denoiser instead, our method improves Top-1 R-precision from 0.555 to 0.585 and reduces MM-Dist from 2.749 to 2.473 on HumanML3D, while Free-T2M retains a lower FID on KIT-ML (0.150 vs.\ 0.182). % Since the two approaches are largely complementary, combining them could be a direction for future work.

\subsection{Qualitative Evaluations}

\begin{figure*}[!htbp]
\includegraphics[width=\textwidth]{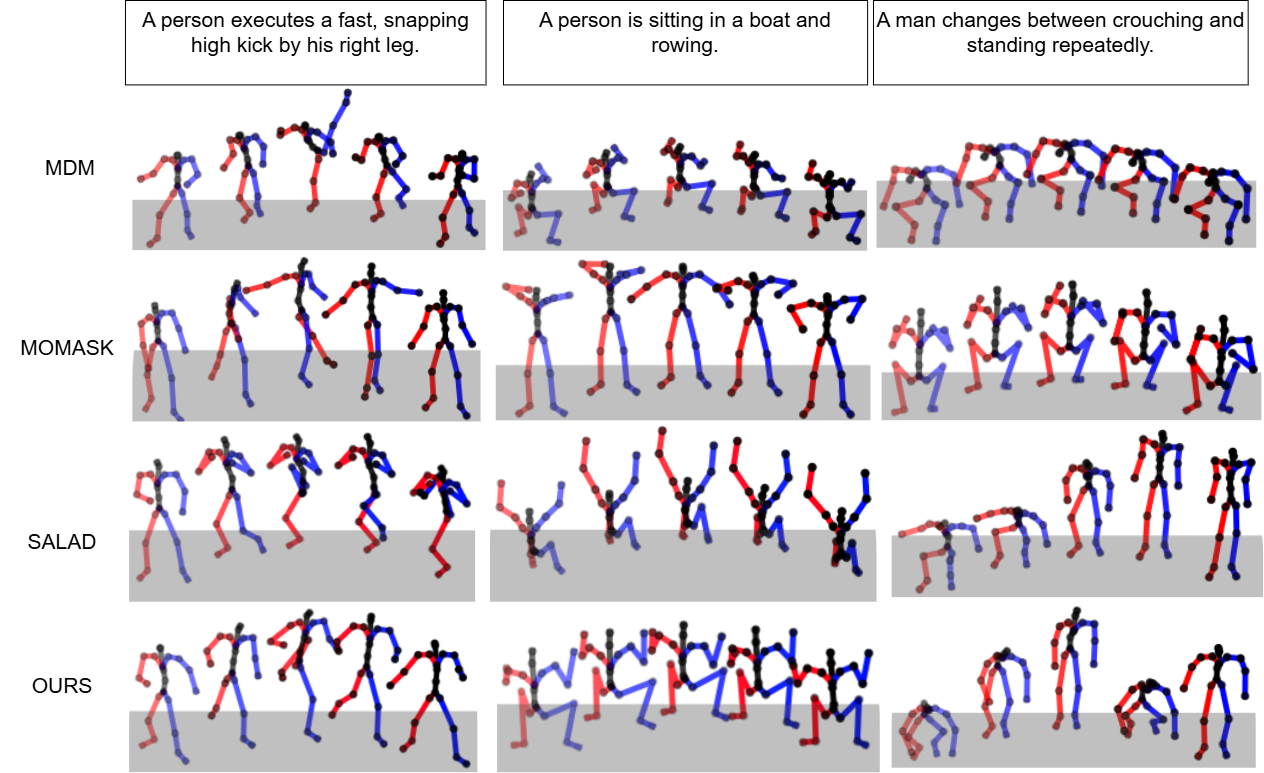}
\caption{Qualitative comparison with state-of-the-art methods, including MDM, MoMask, SALAD, and our method on HumanML3D test dataset.}
\label{fig:qualitative_sota}
\vspace{-5mm}
\end{figure*}
Figure~\ref{fig:qualitative_sota} compares our method with MDM, MoMask, and SALAD on three prompts that differ in both semantic content and temporal dynamics. For ``the fast, snapping high kick'', MoMask keeps the arms extended sideways and produces a less articulated kick, and SALAD drifts sideways before a weak kick, whereas our method moves directly into a clear, high kick with the upper body counterbalancing the movement. This shows that our method is able to incorporate this notion of velocity. For rowing action which includes the constant repetition of a motion, MoMask generates the motion while standing, and SALAD sits but raises the arms overhead; our method combines a seated pose with repeated forward-and-backward arm strokes.  For the alternative motion of crouching and standing (thus periodic change of the action), MDM and MoMask remain crouched with little variation, and SALAD performs only a single transition, whereas our method alternates between the two states several times. 

Overall, our method better reflects both the semantic content of the descriptions, such as the seated posture for rowing, and their temporal dynamics, such as the speed of the kick and the repetition of the crouching and rowing movements. These temporal dynamics are closely related to the frequency content of the motion, which is consistent with the improvements brought by incorporating frequency information into our framework.

\subsection{Ablation Study and Frequency Transform Variants}
\subsubsection{Effect of Frequency Adjustment and Frequency Attention}
\label{sec:ablation_freq}
\begin{table}[!htbp]
\centering
\caption{Ablation study on the frequency components on the test sets of HumanML3D and KIT-ML.}
%  FA and FAttn denote the frequency adjustment in the VAE and the frequency attention in the denoiser, respectively. $\uparrow$ and $\downarrow$ denote higher and lower values are preferable, respectively. \textbf{Bold} and \textcolor{blue}{blue} indicate best and second best results, respectively.
\label{tab:ablation_freq}
% \vspace{-1mm}
\setlength{\tabcolsep}{3pt}
\fontsize{7pt}{9pt}\selectfont
\resizebox{\columnwidth}{!}{%
\begin{tabular}{c c c c c c}
\toprule
 & & \multicolumn{2}{c}{\textbf{HumanML3D}} & \multicolumn{2}{c}{\textbf{KIT-ML}} \\
\cmidrule(lr){3-4} \cmidrule(lr){5-6}
\textbf{FA} & \textbf{FAttn} & R-Precision Top-3 $\uparrow$ & FID $\downarrow$ & R-Precision Top-3 $\uparrow$ & FID $\downarrow$ \\
\midrule
 & & $0.857 \pm 0.002$ & $0.076 \pm 0.002$ & \textbf{0.828 $\pm$ 0.005} & $0.296 \pm 0.012$ \\
\checkmark & & \textcolor{blue}{$0.866 \pm 0.003$} & \textcolor{blue}{$0.055 \pm 0.002$}  & \textcolor{blue}{$0.817 \pm 0.004$} & \textcolor{blue}{$0.200 \pm 0.012$} \\
 & \checkmark & $0.859 \pm 0.002$ & $0.057 \pm 0.002$ & $0.814 \pm 0.004$ & $0.282 \pm 0.012$ \\
\midrule
\checkmark & \checkmark & \textbf{0.868 $\pm$ 0.003} & \textbf{0.052 $\pm$ 0.003}
 & $0.809 \pm 0.006$ & \textbf{0.182 $\pm$ 0.010} \\
\bottomrule
\end{tabular}%
}
% \vspace{-3mm}
\end{table}

Table~\ref{tab:ablation_freq} analyses the contribution of each frequency component. Both frequency modules (frequency enhancement and frequency attention module) improve motion quality on their own, but the frequency adjustment in the VAE (FA) is the main contributor: it reduces FID by 28\% on HumanML3D and 32\% on KIT-ML, and it also raises Top-3 R-precision on HumanML3D from 0.857 to 0.866. This indicates that shaping the frequency content of the latent representation itself provides a better space for the denoiser to model. The frequency attention in the denoiser (FAttn) yields a comparable FID reduction on HumanML3D (0.076 to 0.057) but only a marginal one on KIT-ML, which may be due to the smaller size of KIT-ML being insufficient to train the additional attention layers and the motion jitteriness which can reduce the effectiveness of the attention mechanism (in frequency domain). Combining the two components gives the best FID on both datasets (0.052 and 0.182), showing that exposing frequency information in both the latent representation and the denoiser is complementary. %On KIT-ML, however, each component slightly lowers R-precision, suggesting a modest trade-off between motion fidelity and text alignment on this smaller dataset.

\subsubsection{Choice of Frequency Transform}
\label{sec:ablation_transform}
%, with all other settings unchanged.
\begin{table}[!htbp]
\centering
\caption{Comparison of frequency transforms on the KIT-ML test set. The DCT and SWT is used as frequency transform method.}
\label{tab:ablation_transform}
% \vspace{-1mm}
\setlength{\tabcolsep}{6pt}
% \fontsize{7pt}{9pt}\selectfont
\begin{tabular}{l c c}
\toprule
\textbf{Transform} & FID $\downarrow$ & MM-Dist $\downarrow$ \\
\midrule
SWT & $0.246 \pm 0.011$ & $2.629 \pm 0.020$ \\
DCT & \textbf{0.200 $\pm$ 0.012} & \textbf{2.620 $\pm$ 0.015} \\
\bottomrule
\end{tabular}
\vspace{-5mm}
\end{table}

We adopt the DCT as our frequency transform because of its energy compaction property and the direct control it offers over individual frequency bands. Both of our frequency modules operate at the band level: the frequency enhancement module applies a learnable gain to each band, and the frequency attention treats each band as a token. The DCT provides this representation, with one real coefficient per frequency, ordered from low to high, and it is orthonormal, lightweight, and easily integrated through a single matrix multiplication for sequences of any length. Wavelet transforms (stationary wavelet transform of SWT), in contrast, are designed for multi-scale analysis localized in time. They decompose the signal into a small number of hierarchical sub-bands whose structure depends on the choice of wavelet basis and decomposition depth, which does not map directly onto band-wise gains or band tokens. As shown in Table~\ref{tab:ablation_transform}, directly replacing the DCT with a wavelet transform in our framework leads to lower performance. Exploiting wavelets effectively would require redesigning both components around their hierarchical structure, which can be a potential for the future work.

\section{CONCLUSIONS}
\label{sec:conclusion}

In this work, we address the limitations of current text-to-motion generation methods, which model motion primarily in its raw spatio-temporal form and therefore represent its intrinsic frequency characteristics only implicitly. Building on the skeleton-structured latent space of SALAD method, we introduced frequency information at two stages: a DCT-based frequency enhancement module that refine the encoder feature of each body part through learnable, body-part-specific gains, and a frequency attention mechanism that allows the denoiser to attend across temporal frequency bands. 

Quantitative evaluations on HumanML3D and KIT-ML show that our approach improves our motion generation qualities over the SALAD baseline and achieves competitive performance against state-of-the-art methods, with particularly strong text-motion alignment on HumanML3D dataset. Qualitative and joint-wise spectral analyses show that our method better reproduces the action-specific dynamics described in the text, such as speed and repetition, concentrating spectral power on the relevant body parts.

% Qualitative comparisons shows that our method capable to reproduce the action-specific dynamics described in the text, such as speed, repetition and periodicity. Further analysis of the joint-wise frequency spectra confirms that the generated spectral power concentrates on the body parts and dynamics specified by the prompt. 

Ablation studies show that both components contribute to these improvements, with the frequency enhancement providing the largest gain and the DCT outperforming wavelet transforms, suggesting that structuring frequency information in the latent space is an effective way to improve generation. 

% Ablation studies show that both frequency components contribute to these improvements, with the latent frequency enhancement with DCT based transformation over wavelet based providing the largest improvement to our method. These findings suggest that explicitly structuring frequency information in the latent space is an effective way to improve downstream generation performance. 

Beyond the directions discussed above, future work will explore multi-scale frequency decomposition and adaptive filtering techniques, such as Kalman filtering, to adjust frequency components more
selectively.
 
%%%%%%%%%%%%%%%%%%%%%%%%%%%%%%%%%%%%%%%%%%%%%%%%%%%%%%%%%%%%%%%%%%%%%%%%%%%%%%%%
% \section{ACKNOWLEDGMENTS}

% The authors gratefully acknowledge the contribution of reviewers' comments, etc. (if desired). Put sponsor acknowledgments in the unnumbered footnote on the first page.

%%%%%%%%%%%%%%%%%%%%%%%%%%%%%%%%%%%%%%%%%%%%%%%%%%%%%%%%%%%%%%%%%%%%%%%%%%%%%%%%

%%%%%%%%%%%%%%%%%%%%%%%%%%%%%%%%%%%%%%%%%%%%%%%%%%%%%%%%%%%%%%%%%%%%%%%%%%%%%%%%
% \section*{ETHICAL IMPACT STATEMENT}
% This year, at FG 2027, we will be introducing a new requirement that authors submit an Ethical Impact Statement as part of the submission process. (Note that this requirement applies to all short and long papers submitted to the main track. Each special track may have its own rules.) To support authors in meeting this requirement, as well as reviewers in assessing it, we have prepared this document with general guidelines, a checklist for authors and reviewers to complete, and answers to some frequently asked questions. As this is a new policy, we welcome questions and feedback as we refine it and develop a shared understanding.
%%%%%%%%%%%%%%%%%%%%%%%%%%%%%%%%%%%%%%%%%%%%%%%%%%%%%%%%%%%%%%%%%%%%%%%%%%%%%%%%

{\small
\bibliographystyle{ieee}
\bibliography{egbib}
}

\end{document}